# A Heat-Map-based Algorithm for Recognizing Group Activities in Videos

Weiyao Lin, Hang Chu, Jianxin Wu, Bin Sheng, and Zhenzhong Chen

**Abstract**— **In this paper, a new heat-map-based (HMB) algorithm is proposed for group activity recognition. The proposed algorithm first models human trajectories as series of "heat sources" and then applies a thermal diffusion process to create a heat map (HM) for representing the group activities. Based on this heat map, a new key-point based (KPB) method is used for handling the alignments among heat maps with different scales and rotations. And a surface-fitting (SF) method is also proposed for recognizing group activities. Our proposed HM feature can efficiently embed the temporal motion information of the group activities while the proposed KPB and SF methods can effectively utilize the characteristics of the heat map for activity recognition. Experimental results demonstrate the effectiveness of our proposed algorithms.**

## I. INTRODUCTION

Detecting group activities or human interactions has attracted increasing research interests in many applications such as video surveillance and human-computer interaction [1-6].

Many algorithms have been proposed for recognizing group activities or interactions [1-6, 24-25]. Zhou et al. [2] propose to detect pair-activities by extracting the causality, mean, variance features from bi-trajectories. Ni et al. [3] further extend the causality features into three types of individuals, pairs and groups. Besides, Chen et al. [5] detect group activities by introducing the connected active segmentations for representing the connectivity among people. Cheng et al. [4] propose the Group Activity Pattern for representing group activities as Gaussian processes and extract Gaussian parameters as features. However, most of the existing algorithms extract the overall features from the activities' entire motion information (e.g., the statistical average of the motion trajectory). These features cannot suitably embed activities' temporal motion information (e.g., fail to indicate where a person is in the video at a certain moment). Thus, they will have limitations when recognizing more complex group activities. Although some methods [6, 29] incorporate the temporal information with chain models such as the Hidden Markov Models (HMM), they have the disadvantage of requiring large-scale training data [17]. Besides, other methods try to include the temporal information by attaching time stamps with trajectories and perform recognition by associating these time stamp labels [18-19]. However, these methods are more suitable for scenarios with only one trajectory or trajectories with fixed correspondence. They will become less effective or even infeasible when describing and differentiating the complicated temporal interactions among multiple trajectories in group activities. Furthermore, [24] and [25] give more extensive survey about the existing techniques used in group activity recognition and crowd analysis.

In another part, handling motion uncertainties is also an important issue in group activity recognition. Since the motions of people vary inherently in group activities, the recognition accuracy may be greatly affected by this uncertain motion nature. Although some methods utilize Gaussian processes estimation or filtering to handle this uncertain problem [3, 4], they do not simultaneously consider the issue for reserving the activities' temporal motion information.

Furthermore, the recognition method is a third key issue for recognizing group activities. Although the popularly-used models such as Linear Discriminative Analysis and HMM [6] show good results in many scenarios, their training difficulty and the requirement of the training data scale will increase substantially when the feature vector length becomes large or the group activity becomes complex. Therefore, it is also non-trivial to develop more flexible recognition methods for effectively handling the recognition task.

In this paper, we propose a new heat-map-based (HMB) algorithm for group activity recognition. The contributions of our work can be summarized as follows:

(1) We propose a new heat map (HM) feature to represent group activities. The proposed HM can effectively catch the temporal motion information of the group activities.

(2) We propose to introduce a thermal diffusion process to create the heat map. By this way, the motion uncertainty from different people can be efficiently addressed.

(3) We propose a key-point based (KPB) method to handle the alignments among heat maps with different scales and rotations. By this way, the heat map differences due to motion uncertainty can be further reduced and the follow-up recognition process can be greatly facilitated.

(4) We also propose a new surface-fitting (SF) method to recognize the group activities. The proposed SF method can effectively catch the characteristics of our heat map feature and perform recognition efficiently.

The remainder of this paper is organized as follows. Section II describes the basic ideas of our proposed HM feature as well as the KPB and SF methods. Section III presents the details of our HMB algorithm. The experimental results are shown in Section IV and Section V concludes the paper.

## II. BASIC IDEAS

### A. The heat-map feature

As mentioned, given the activities' motion information (i.e., motion trajectory in this paper), directly extracting the

W. Lin and H. Chu are with the Department of Electronic Engineering, Shanghai Jiao Tong University, Shanghai 200240, China (email: {wylin, chuang321}@sjtu.edu.cn).

J. Wu is with the Department of Computer Science, Nanjing University, China (email: wujx@lamda.nju.edu.cn).

B. Sheng is with the Department of Computer Science, Shanghai Jiao Tong University, Shanghai 200240, China (email: shengbin@cs.sjtu.edu.cn).

Z. Chen is with MediaTek Inc., USA (zzchen@ieee.org).

global features will lose the useful temporal information. In order to avoid such information loss, we propose to model the activity trajectory as a series of heat sources. As shown in Figure 1, (a) is the trajectory of one person. In order to transfer the trajectories into heat source series, we first divide the entire video scene into small non-overlapping patches (i.e., the small squares in (b)). If the trajectory goes through a patch, this patch will be defined as one heat source. By this way, a trajectory can be transferred into a series of heat sources, as in Figure 1 (b). Furthermore, in order to catch the temporal information of the trajectory, we also introduce a decay factor on different heat sources such that the thermal energies of the "older" heat sources (i.e., patches closer to the stating point of the trajectory) are smaller while the "newer" heat sources will have larger thermal energies. By this way, the thermal values of the heat source series can be arranged increasingly according to the direction of the trajectory and the temporal information can be effectively embedded.

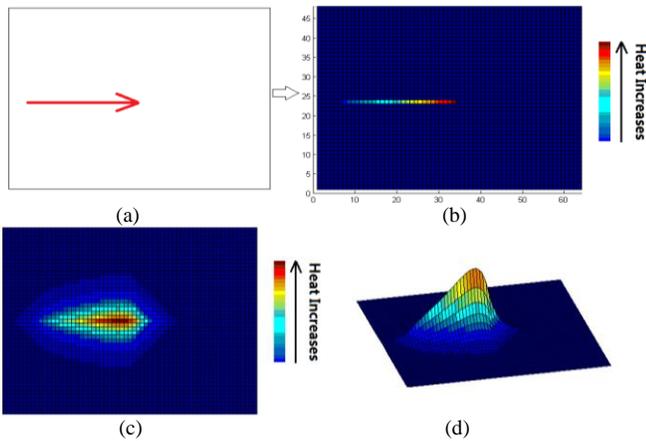

Figure 1. (a): The activity trajectory; (b) The corresponding heat source series; (c) The heat map (HM) diffused from the heat source series in (b); (d) The HM surface of (c) in 3D.

Finally, since people's trajectories may have large variations, directly using the heat source series as features will be greatly affected by this motion fluctuation. Therefore, in order to reduce the motion fluctuation, we further propose to introduce a thermal diffusion process to diffuse the heats from the heat source series to the entire scene. We call this diffusion result as the heat map (HM). With our HM feature, we can describe the activities' motion information by 3D surfaces. Figure 1 (c) and (d) show the HM of the trajectory in Figure 1 (a) in 2D format and in 3D surface format, respectively. Several points need to be mentioned about the HM in our paper:

(1) Note that although the heat diffusion was introduced in object segmentation in some works [8], the mechanism and utilization of HM in our algorithm is far different from them. And to the best of our knowledge, this is the first work to introduce HM into group activity recognition.

(2) The definition of "heat map" in this paper is also different from the ones used in some activity recognition methods [11-12]. In those methods [11-12], the heat maps are defined to reflect the number of translations among different regions without considering the order during passes. Thus, they are more focused on reflecting the "popularity" of regions (i.e., whether some regions are more often visited by people) while neglecting the temporal motion information as well as the interactions among trajectories.

(3) With the HM features, we can perform offline activity recognition by creating HMs for the entire trajectories. This off-line recognition is important in many applications such as video retrieval and surveillance video investigation [6, 15]. Furthermore, the HM features can also be used to perform on-line (or on-the-fly) recognition by using shorter sliding windows. This point will be further discussed in Section IV.

After the calculation of HM features, we can use them for recognizing group activities. However, two problems need to be solved for perform recognition with HM features. They are described in the following.

### B. The alignments among heat maps

Although the thermal diffusion process can reduce the motion fluctuation effect due to motion uncertainty or tracking biases, the resulting HM will still differ a lot due to the various motion patterns for different activities. For example, in Figure 2 (a), since the trajectories of human activities take varying directions and lengths, the heat maps for the same type of group activity show large differences in scales and rotations. Therefore, alignments are necessary to reduce these HM differences for facilitating the follow-up recognition process.

In this paper, we propose a new key-point based (KPB) method to handle the alignments among heat maps. Since our heat maps are featured with peaks (i.e., local maxima in HM as in Figure 1 (d)), the proposed KPB method extracts the peaks from HMs as the key points and then performs alignments according to these key points in an iterative manner. By this way, the scale and rotation variations among heat maps can be effectively removed. Figure 2 (b) shows the alignment results of the heat maps in (a) by our KPB method. More details about the KPB method will be described in the next section.

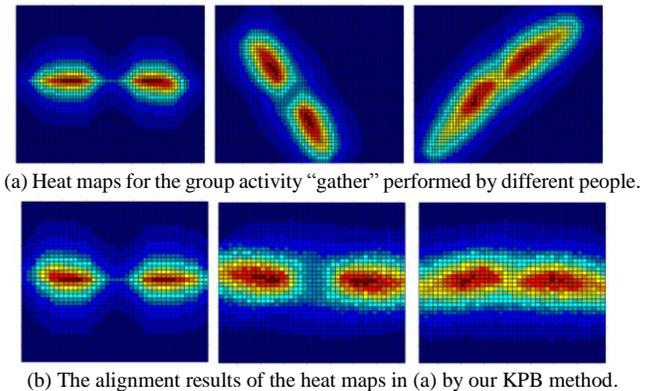

(a) Heat maps for the group activity "gather" performed by different people.

(b) The alignment results of the heat maps in (a) by our KPB method.
Figure 2. The alignments among heat maps.

### C. Recognition based on the heat maps

Since the HM feature includes rich information, the problem then comes to the selection of a suitable method for performing recognition based on this HM feature. In this paper, we further propose a surface-fitting (SF) method for activity recognition. In our SF method, a set of standard surfaces are first identified for representing different activities. Then, the similarities between the surface of the input HM and the standard surfaces are calculated. And finally, the best matched standard surface will be picked up and its corresponding activity will become the recognized activity for the input HM. The process of our SF method is shown in Figure 3.

With the basic ideas of the HM feature, the KPB and the SF methods described above, we can propose our heat-map-based (HMB) group activity recognition algorithm. It is described in detail in the following section.

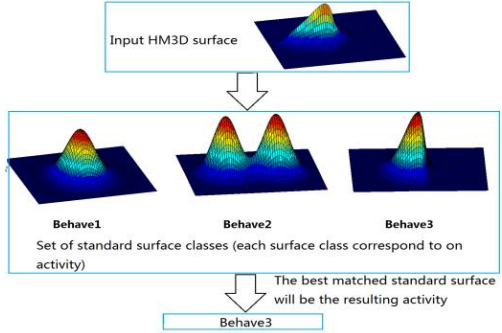

Figure 3. The process of the surface-fitting (SF) method.

## III. THE HMB ALGORITHM

The framework of our HMB algorithm can be described by Figure 4. In Figure 4, the input group activities' trajectories are first transferred into heat source series, then the thermal diffusion process is performed to create the HM feature for describing the input group activity. After that, the KPB method is used for aligning HMs and finally the SF method is used for recognizing the group activities. As mentioned, the heat source series transfer, the thermal diffusion, the KPB method, and the SF method are the four major contributions of our proposed algorithm. Thus, we will focus on describing these four parts in the following.

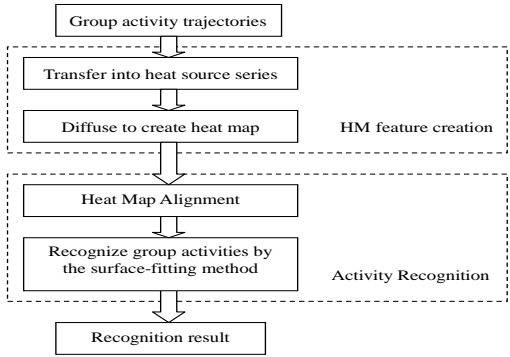

Figure 4. The process of the HMB algorithm.

### A. Heat source series transfer

Assume that we have in total $j$ trajectories in the current group activity. The thermal energy $E_i$ of the heat source patch $i$ can be calculated by:

$$E_i = \sum_j \overline{E_{i,j}} \cdot e^{-k_t\left(t_{cur}-t_{id,j}\right)} \qquad (1)$$

where $e^{-k_t\left(t_{cur}-t_{id,j}\right)}$ is the time decay term [10], $k_t$ is the temporal decay coefficient, $t_{cur}$ is the current frame number, and $t_{id,j}$ is the frame number when the $j$-th trajectory leaves patch $i$.

$\overline{E_{i,j}}$ is the accumulated thermal energy for trajectory $j$ in patch $i$ and it can be calculated by Eq. (2). From Eq. (1), we can see that "newer" heat sources of the trajectory have more thermal energies than the "older" heat sources.

$$\overline{E_{i,j}} = \int_{0}^{t_{id,j}-t_{is,j}} C \cdot e^{-k_t t} dt = \frac{C}{k_t}\left(1 - e^{-k_t\left(t_{id,j}-t_{is,j}\right)}\right) \qquad (2)$$

where $t_{is,j}$ and $t_{id,j}$ are the frame number when the $j$-th trajectory enters and leaves patch $i$, respectively. $k_t$ is the temporal decay coefficient as in Eq. (1), and $C$ is a constant. In the experiments of our paper, $C$ is set to be 1. From Eq. (2), we can see that the accumulated thermal energy is proportional to the stay length of trajectory $j$ at patch $i$. If $j$ stays in $i$ for longer time, more thermal energy will be accumulated in patch $i$. On the other hand, if no trajectory goes through patch $i$, the accumulated thermal energy of patch $i$ will be 0, indicating that patch $i$ is not a heat source patch.

### B. Thermal diffusion

After getting the heat source series by Eq. (1), the thermal diffusion process will be performed over the entire scene to create the HM. The HM value $H_i$ at patch $i$ after diffusion [10] can be calculated by:

$$H_i = \frac{\sum_{l=1}^{N} E_l \cdot e^{-k_p d(i,l)}}{N} \qquad (3)$$

where $E_l$ is the thermal energy of the heat source patch $l$, $N$ is the total number of heat source patches. $k_p$ is the spatial diffusion coefficient, and $d(i, l)$ is the distance between patches $i$ and $l$.

The advantage of the thermal diffusion process can be described by Figure 5. In Figure 5, the left column lists two different trajectory sets for the group activity "gather". Due to the variation of human activity or tracking biases, these two trajectory sets are obviously different from each other. And these differences are exactly transferred to their heat source series (the middle column). However, with the thermal diffusion process, the trajectory differences are suitably "blurred", which makes their HMs (the right column) close to each other. At the same time, the temporal information of the two group activities is still effectively reserved in the HMs.

Also, Figure 6 shows the example HM surfaces for different group activities defined in Table 1. From Figure 6, it is clear that our proposed HM can precisely catch the activities' temporal information and show obviously distinguishable patterns among different activities.

Furthermore, it should be noted that our proposed HMB algorithm is not limited to trajectories. More generally, as long as we can detect patches with motions, we can use these motion patches as the heat sources to create heat maps. Therefore, in practice, when reliable trajectories cannot be achieved, we can even skip the tracking process and use various low-level motion features (such as the optical flow [28]) to create the heat maps for recognition. This point will be further demonstrated in Section IV

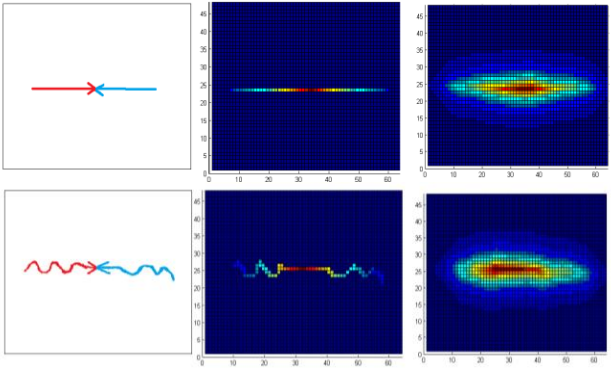

Figure 5. Left column: two trajectory sets for group activity "Gather"; Middle column: the corresponding heat source series; Right column: the corresponding heat maps.

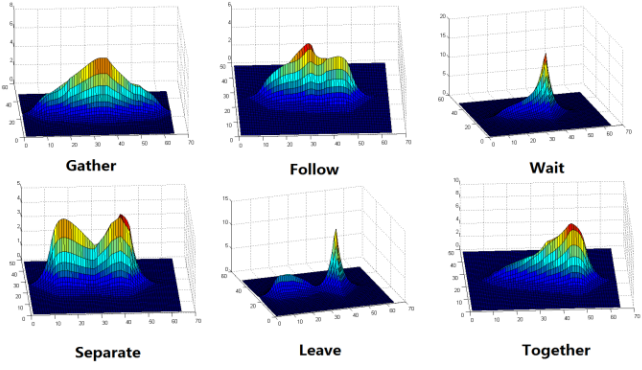

**Gather**    **Follow**    **Wait**

**Separate**    **Leave**    **Together**

Figure 6. The HM surfaces for different group activities.

Table 1 Definitions of different human group activities

| Gather | Two or more persons are gathering to a point. |
|---|---|
| Follow | One or one group of person is followed by one person. |
| Wait | One or one group of person is waiting for one person. |
| Separate | Two or more persons are separating from each other. |
| Leave | One person is leaving one or one group of unmoving person. |
| Together | Two or more persons are walking together. |

### C. The key-point based (KPB) HM alignment method

After generating the HM features, the alignment process is performed for removing the scale and rotation variations among HMs. In this paper, we borrow the idea of the active appearance model (AAM) used in face fitting [7, 13] and propose a key-point-based (KPB) HM alignment method. The process of using our KPB method to align an input HM with a target HM can be described by Algorithm 1.

Furthermore, several points need to be mentioned about our KPB method:

(1) When HMs with different peak numbers are aligned, only the peaks available in both HMs are used for alignment (e.g., when an HM with $n_1$ peaks is aligning with an HM with $n_2$ peaks and $n_1 < n_2$, we only use $n_1$ peaks as the key points for alignment).

(2) For HM with only one peak, we will add an additional key point for alignment. That is, we first pick up the points whose heat values are half of that of the peak point, and then the one which is farthest to the peak will be selected as the second key point for alignment, as shown in Figure 7. Since the direction from the peak to the additional key point represents the slowest-descending slope of the HM surface, the HMs can then be suitably aligned by matching this slope.

(3) It should be noted that in the steps 2, 3, and 4 in Algorithm 1, the key points are shifted, scaled, and rotated coherently (i.e., by the same parameter) in order to keep the overall shape of HM during alignment.

(4) In Algorithm 1, the key points $G_i$ of the target HM are assumed to be already shifted and scaled properly. In our HMB algorithm, we perform clustering on the HMs in the training data and perform alignment within each cluster. After that, the mean of the aligned HMs in each cluster is used as the standard surface (i.e., the target HM) for representing the cluster during recognition. The process of clustering the HMs and calculating the mean HM for each cluster is performed in an iterative manner as described by Algorithm 2. And this point will be further discussed in detail in the next section.

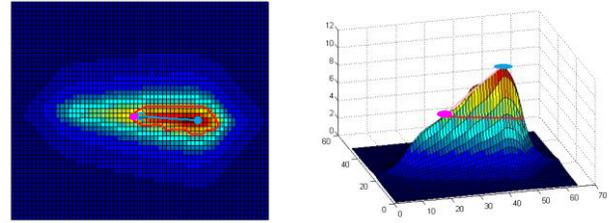

(a) The HM in 2d view    (b) The HM of (a) in 3D view

Figure 7. The selection of the second key point for single-peak HM cases (the pink point is the selected second key point, the blue point is the peak point, and the red line is the contour line whose corresponding point values are all equal to the half of the peak value, best view in color).

---

**Algorithm 1** The KPB Method

1. For the input HM, extract the $n$ largest peak points and use the locations of these peak points as the key points in later alignment steps: ($P_1$, $P_2$, $P_3$, ... $P_n$), where $P_i=[x_i, y_i]$ is the location of the $i$-th key point with $x_i$ and $y_i$ being its $x$ and $y$ coordinates in the HM.

2. Organize the key points $P_i$ for each input HM in an descending order according to their heat values in HM (i.e., $H(P_i) > H(P_j)$ for $i > j$).

3. Shift the key points ($P_1$, $P_2$, $P_3$, ... $P_n$) such that the gravity center of these points is in the center of the HM.

4. Scale the key points ($P_1$, $P_2$, $P_3$, ... $P_n$) such that $\left( \sum_{i=1}^{n} \sqrt{x_i^2 + y_i^2} \right)/n = 1$.

5. Align the key points of the input HM with the target HM such that $\arg_T \left( \min \left( \sum_{i=1}^{n} |G_i - P_i \cdot T|^2 \right) \right)$, where $T$ is a 2×2 matrix for aligning the key points of $P_i$, and $G_i$ are the key points for the target HM. $T$ can be achieved by linear regression.

6. Apply the final shift, rotation, and scaling operation derived from 2-4 on the entire input HM for achieving the final aligned version.

---

### D. The surface-fitting (SF) method for activity recognition

With the HM feature and the KPB alignment method, we can then perform recognition based on our surface-fitting (SF) method. The surface-fitting process can be described by Eq. (4):

$$m^* = \arg \min_{m} \left( \min_{T_m} // T_m \cdot S_{HM} - S_{SD,m} // \right) \quad (4)$$

where $m^*$ is the final recognized activity. $S_{HM}$ is the HM surface of the input activity, $S_{SD,m}$ is the standard surface for activity $m$. $T_m$ is the alignment operator derived by Algorithm 1 for aligning with $S_{SD,m}$. And $\| \cdot \|$ is the absolute difference between two HM surfaces. From Eq. (4), we can see that the SF method includes two steps. In the first step, the input HM is aligned to fit with each standard surface. And then in the second step, the standard surface that best fits the input HM surface will be selected as the recognized activity.

---

**Algorithm 2** Clustering the HMs and calculating the mean HM key points for each cluster in the training set

---

1. Cluster the HMs in the training set according to their activity labels.
2. **for** each HM $v$ in the training set **do**
3.     Shift the key points ($P_{1,v}$, $P_{2,v}$, $P_{3,v}$, ... $P_{n,v}$) of HM $v$ such that the gravity center of these point is in the center of the HM.
4.     Scale the key points ($P_{1,v}$, $P_{2,v}$, $P_{3,v}$, ... $P_{n,v}$) of HM $v$ such that $\left( \sum_{i=1}^{n} \sqrt{x_{i,v}^2 + y_{i,v}^2} \right)/n = 1$.
5. **end for**
6. **for** each cluster $u$ **do**
7.     Randomly select an HM in cluster $u$ as the initial mean HM and define the key points for this mean HM as ($G_1^u$, $G_2^u$, $G_3^u$, ... $G_n^u$)
8.     **for** each HM $v$ in cluster $u$ **do**
9.         Scale the key points ($P_{1,v}$, $P_{2,v}$, $P_{3,v}$, ... $P_{n,v}$) of HM $v$ such that $\left( \sum_{i=1}^{n} \sqrt{x_{i,v}^2 + y_{i,v}^2} \right)/n = 1$.
10.         Align the key points of the HM $v$ with the current mean HM such that $\arg_{T_v} \left( \min \left( \sum_{i=1}^{n} | G_i^u - P_{i,v} \cdot T_v |^2 \right) \right)$, where $T_v$ is the alignment matrix for $v$.
11.         Move the key points of the HM $v$ to the aligned places, i.e., $P_{i,v}^{new} = P_{i,v} \cdot T_v$.
12.     **end for**
13.     Update the key points of the mean HM of cluster $u$ by: $G_i^{u,new} = \left( \sum_{v=1}^{NUM} P_{i,v}^{new} \right)/NUM$, where $NUM$ is the number of HMs in cluster $u$.
14.     If not converged and iteration time$\leq$ 1000, return to 8.
15.     Align all the HMs in cluster $u$ to the calculated key points of the mean HM. And the final mean HM can be achieved by averaging or selecting the most fitted one among these aligned HMs.
16. **end for**

---

As shown in Algorithm 2, the standard surface can be achieved by clustering the training HMs and taking the mean HM for each cluster. However, since the HMs may still vary within the same activity, it may still be less effective to use one fixed HM as the standard surface for recognition. Therefore, in this paper, we further propose an adaptive surface-fitting (ASF) method which selects the standard surface in an adaptive way. The proposed ASF method can be described by Eq. (5):

$$m^* = arg\,max_m \left( \sum_{S_{tr,m} \in N_w(S_{HM})} GA \left( \min_{T_v} \left\| T_{tr} \cdot S_{HM} - S_{tr,m} \right\| \right) \right) \quad (5)$$

where $S_{HM}$ is the HM surface for the input activity. $S_{tr,m}$ is the HM surface for activity $m$ in the training data. $T_{tr}$ is the alignment operator for aligning with $S_{tr}$. $N_w(S_{HM})$ is set containing the $w$ most similar HM surfaces to $S_{HM}$. $GA(\cdot)$ is the Gaussian kernel function as defined by Eq. (6):

$$GA(\mathbf{x}) = exp\left( -\frac{|\mathbf{x}|^2}{2\sigma^2} \right) \quad (6)$$

where $\sigma$ controls the steepness of the kernel.

From Eq. (5), we can see that the proposed ASF method adaptively select the most similar HM surfaces as the standard surfaces for recognition. By this way, the in-class HM surface variation effect can be effectively reduced. Furthermore, by introducing the Gaussian kernel, different training surfaces $S_{tr,m}$ can be allocated different importance weights according to their similarity to the input HM $S_{HM}$ during the recognition process.

Furthermore, several things need to be mentioned about the ASF method:

(1) When $w > 1$ in $N_{w,m}(S_{HM})$, the ASF method can be viewed as an extended version of the $k$-nearest-neighbor (KNN) methods [14] where the kernel-weighted distance between points is calculated by the absolution difference between the aligned HM surfaces.

(2) When $w = 1$ in $N_{w,m}(S_{HM})$, the ASF method is simplified to finding a $S_{tr,m}$ in the training set that can best represent the input $S_{HM}$.

## IV. EXPERIMENTAL RESULTS

In this section, we show experimental results for our proposed HMB algorithm. The ASF method is used for recognition in our experiments. And the patch size is set to be $10 \times 10$ based on our experimental statistics, in order to achieve satisfactory resolution of the HM surface while maintaining the efficiency of computation. Furthermore, for each input video clip, the heat map is created for the entire clip.

### A. Experimental results on the BEHAVE dataset

In this sub-section, we perform five different sets of experiments on the BEHAVE dataset to evaluate our proposed algorithm.

First of all, we change the values of the temporal decay parameter $k_t$ and the thermal diffusion parameter $k_p$ in Eqs (1) and (3) to see their effects in recognition performances. We select 200 video clips from the BEHAVE dataset [1] and recognize six group activities defined in Table 1. The sample number for each group activity is shown in Table 2. Each video clip includes 2-5 trajectories. In order to examine the algorithm's performance against tracking fluctuation and tracking bias, we perform 5 rounds of experiments where in each round, different fluctuation and biases effects are added on the ground-truth trajectories. The final results are averaged over the five rounds. The recognition results under 75%-training and 25%-testing are shown in Tables 3 and 4.

Table 2 The sample number for different group activities for the experiments in Tables 3-4 and Figures 8-9

| Gather | Follow | Wait | Separate | Leave | Together | **Total** |
|--------|--------|------|----------|-------|----------|-----------|
| 33 | 33 | 34 | 33 | 33 | 34 | **200** |

Table 3 The TER rates of HMB algorithm under different spatial diffusion coefficient $k_p$ values (when $k_t$ =0.125)

| | $k_p$=0.1 | $k_p$ =1 | $k_p$ =2 | $k_p$ =3 | $k_p$ =4 | $k_p$ =5 | $k_p$ =$\infty$ |
|--|-----------|----------|----------|----------|----------|----------|------------------|
| HMB ($w$=1) | 42% | 16% | 13% | 15% | 17% | 19% | 38% |

Table 4 TER rates of HMB algorithm under different temporal diffusion coefficient $k_t$ values (when $k_p$ =2)

| | $k_t$=0.025 | $k_t$=0.125 | $k_t$=0.25 | $k_t$=0.5 | $k_t$=1 | $k_t$=∞ |
|---|---|---|---|---|---|---|
| HMB ($w$=1) | 25% | 13% | 15% | 17% | 20% | 45% |

Tables 3-4 show the total error rate (TER) for different $k_t$ and $k_p$ values. The TER rate is calculated by $N_{t\_miss}/N_{t\_f}$ where $N_{t\_miss}$ is the total number of misdetection activities for both normal and abnormal activities and $N_{t\_f}$ is the total number of activity sequences in the test set [6, 15]. TER reflects the overall performance of the algorithm in recognizing all activity types [6, 15]. In Tables 3-4, our HMB algorithm is performed where $w$ in Eq. (5) is set to be 1 (i.e., selecting only the most similar HM surface in the training set during recognition). Furthermore, the example HM surfaces under different $k_t$ and $k_p$ values are shown in Figures 8 and 9, respectively.

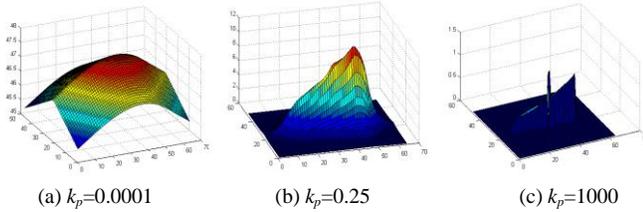

(a) $k_p$=0.0001    (b) $k_p$=0.25    (c) $k_p$=1000
Figure 8. Example HM surfaces for activity "together" with different $k_p$ values.

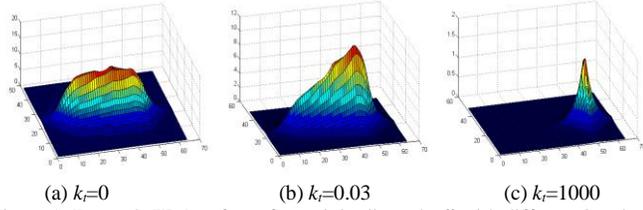

(a) $k_t$=0    (b) $k_t$=0.03    (c) $k_t$=1000
Figure 9. Example HM surfaces for activity "together" with different $k_t$ values.

From Table 3 and Figure 8, we can see that: (1) When $k_p$ is set to be a very small number (such as Figure 8 (a)), the thermal diffusion effect is too strong that the HM is close to a flat surface. In this case, the effectiveness of the HM cannot fully work and the recognition performances will be decreased. (2) On the contrary, if $k_p$ is set to be extremely large (such as Figure 8 (c)), few thermal diffusion is performed and the HM surfaces are only concentrated on the heat source patches. In these cases, the recognition performance will also reduce. (3) The results for $k_p$ =2 and $k_p$ =∞ in Table 3 can also show the usefulness of our proposed thermal diffusion process. Since no diffusion process is applied on the HM when $k_p$ =∞, it is more vulnerable to tracking fluctuation or tracking biases, resulting in lower recognition results. Comparatively, by the introduction of our thermal diffusion process, the tracking fluctuation effects can be greatly reduced and the performances can be obviously improved. (4) If taking a careful look at Figure 8 (c), we can see that there is a needle-like peak in the middle of the HM. It is created because both trajectories traverse the same patch, thus making the heat source value greatly amplified at this patch location. If we directly use this HM for recognition, this "noisy" peak will affect the final performance. However, by using our heat diffusion process, this noisy peak can be blurred or deleted (such as Figure 8 (b)) and the coherence among HMs can be effectively kept. (5) Except for extremely small or large values, $k_p$ can achieve good results within a wide range.

From Table 4 and Figure 9, we can see the effects of the temporal decay parameter $k_t$. (1) For an extremely small $k_t$ value (such as Figure 9 (a)), most heat sources will show the same values. In this case, the temporal information of the trajectory will be lost in the HM and the performance will be reduced. (2) For an extremely large $k_t$ value (such as Figure 9 (c)), the "old" heat sources will decay quickly such that the HM is only concentrated on the "newest" hear source. In this case, the trajectory's temporal information will also be lost and leading to low performances. (3) Except for extremely small or large values, $k_t$ can also achieve good results within a wide range.

Based on the above discussion, $k_t$ and $k_p$ in Eqs (1) and (3) are set to be 0.125 and 2 respectively throughout our experiments.

Secondly, we compare our HMB algorithm with the other algorithms. In order to include more activity samples, we further increase the sample number and select 325 video clips for six activities (as in Table 1) from the BEHAVE dataset [1]. The sample number distributions for different activities are shown in Table 5. Each video clip includes 2-5 trajectories. Figure 10 show some examples of the six activities. The following 6 algorithms are compared:

(1) The WF-SVM algorithm which utilizes causalities between trajectories for group recognition [2] (WF-SVM).

(2) The LC-SVM algorithm which includes the individual, pair, and group correlations for recognition [3] (LC-SVM).

(3) The GRAD algorithm which uses Markov chain models for modeling the temporal information for performing recognition [6] (GRAD).

(4) Using our proposed HM as the input features and our KPB method for HM alignments. After that, using Principle Component Analysis (PCA) for reducing the HM feature vector length and use Support Vector Machine (SVM) for activity recognition [16, 17] (HM-PCASVM).

(5) Using the entire version of our proposed HMB algorithm and $w$ in Eq. (5) is set to be 1 (HMB($w$=1)).

(6) Using the entire version of our proposed HMB algorithm and $w$ in Eq. (5) is set to be 3 (HMB($w$=3)).

Table 5 The video-clip number for different group activities for the experiments in Tables 6-7

| Gather | Follow | Wait | Separate | Leave | Together | **Total** |
|---|---|---|---|---|---|---|
| 45 | 40 | 76 | 40 | 58 | 66 | **325** |

Similarly, we split the dataset into 75% training-25% testing parts and perform recognition on the testing part [6]. Six independent experiments are performed and the results are averaged. Furthermore, we use the ground-truth trajectories in this experiment. However, note that in practice, various object detection and tracking algorithms [9, 26, 30, 31] can be utilized to achieve trajectories. And even in cases when reliable trajectories cannot be achieved, other low-level features [28] can be used in an algorithm to take the place of the trajectories. This point will be further discussed in Section IV-C later. Table 6 shows the Miss, False Alarm (FA), and Total Error Rates (TER) [6] for different algorithms. The miss detection rate is defined by $N_\theta^{fn}/N_\theta^+$ where $N_\theta^{fn}$ is the number of false negative (misdetection) sequences for activity $\theta$, and $N_\theta^+$ is the total number of positive sequences of activity $\theta$ in the test data [6, 15]. And the FA rate is defined by $N_\theta^{fp}/N_\theta^-$ where $N_\theta^{fp}$ is the number of false positive (false alarm) video clips for activity $\theta$, and $N_\theta^-$ is the total number of negative video clips except activity $\theta$ in the test data [6].

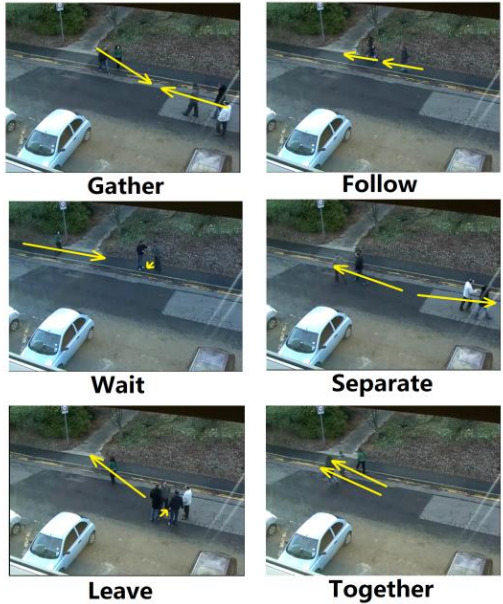

Figure 10. Examples of human group activities in the BEHAVE dataset [1].

From Table 6, we can have the following observations:

(1) Due to the complexity and uncertainty of human activities, the WF-SVM, LC-SVM, and GRAD algorithms still produce unsatisfactory results for some group activities such as "Gather". Compared to these algorithms, algorithms based on our HM features (HMB ($w$=1), HMB ($w$=3), and HM-PCASVM) have better performances. This demonstrates that our HM features are able to precisely catch the characteristics of activities.

(2) Comparing the HMB algorithms (HMB ($w$=1), HMB ($w$=3)) and the HM-PCASVM, we can see that the HMB algorithms have improved results than that of the HM-PCASVM algorithm. This demonstrates the effectiveness of our surface-fitting recognition methods. Note that the improvement of our HMB algorithm will become more obvious in another dataset, as will be shown later.

(3) The performance of the HMB ($w$=1) algorithm is close to the HMB ($w$=3) algorithm. And similar observations can be achieved for other datasets and for other $w$ values (when $w < 5$). Therefore, in practice, we can simply set $w$=1 when implementing the ASF methods.

Thirdly, in order to evaluate the influence of trajectory qualities to the algorithm performances, we perform another experiment by adding Gaussian noises with different strength on the ground-truth trajectories and perform recognition on these "noisy" trajectories. The results are shown in Table 7.

Table 6 The Miss, FA, and TER rates for different algorithms on the BEHAVE dataset

| | | HMB ($w$=3) | HMB ($w$=1) | HM-PCASVM | WF-SVM [2] | LC-SVM [3] | GRAD [6] |
|---|---|---|---|---|---|---|---|
| Gather | Miss | 6.9% | 7.1% | 6.7% | 12.0% | 22.2% | 11.3% |
| | FA | 0.0% | 0.7% | 1.4% | 0.3% | 4.3% | 1.6% |
| Follow | Miss | 0.2% | 2.5% | 6.4% | 8.3% | 17.5% | 16.2% |
| | FA | 0.4% | 0.7% | 1.1% | 2.3% | 2.9% | 1.3% |
| Wait | Miss | 4.3% | 2.6% | 7.2% | 9.0% | 22.4% | 14.3% |
| | FA | 0.8% | 1.1% | 0.4% | 2.0% | 6.4% | 1.8% |
| Separate | Miss | 0.0% | 0.0% | 0.1% | 5.0% | 7.5% | 8.2% |
| | FA | 0.1% | 0.3% | 0.2% | 0.4% | 1.4% | 1.0 % |
| Leave | Miss | 2.9% | 4.1% | 3.0% | 4.8% | 15.2% | 9.6% |
| | FA | 1.5% | 0.8% | 1.9% | 1.4% | 1.5% | 2.5% |
| Together | Miss | 3.5% | 3.9% | 3.7% | 2.9% | 3.5% | 2.6% |
| | FA | 1.9% | 2.3% | 1.5% | 1.8% | 1.9% | 0.8% |
| TER | | 3.6% | 4.0% | 5.2% | 7.0% | 13.9% | 10.9% |

Table 7 Comparison of TER rates with different trajectory qualities ($m$ is the noise strength parameter which measures the average pixel-level deviation from the ground-truth trajectories).

| | $m$=0 | $m$=1 | $m$=2 | $m$=3 | $m$=4 | $m$=5 | $m$=25 |
|---|---|---|---|---|---|---|---|
| HMB($w$=3) | 3.6% | 3.4% | 4.0% | 3.7% | 4.3% | 4.3% | 10.8% |
| WF-SVM | 7.0% | 7.2% | 8.3% | 9.7% | 10.4% | 10.7% | 15.1% |

Table 7 compares the Total Error Rates (TER) of our HMB algorithm and the WF-SVM algorithm [2]. We select to compare with WF-SVM because it has the best performance among the compared methods in Table 6. The noise strength parameter $m$ in Table 7 is the average pixel-level deviation from the ground-truth trajectory. For example, $m$=5 means that in average, the noisy trajectory is 5-pixel deviated from the ground truth trajectory. Note that $m$ only reflects the "average" deviation while the actual noisy trajectories may have more fluctuation effects, for example, fluctuating with different deviation strength around the ground-truth trajectory and deviating with large magnitudes from the ground-truth.

From Table 7, we can see that:

(1) Our HMB algorithm can still achieve pretty stable performances when the qualities of the trajectories decrease (i.e., when the noise strength $m$ increases). Comparatively, the performance decrease by the WF-SVM algorithm is more obvious. For example, when $m$=5, the TER rate of WF-SVM will be increased by more than 3% while our HMB is only increased by less than 1%. This further demonstrates that the heat thermal diffusion process in our algorithm can effectively reduce the possible trajectory fluctuations.

(2) When the noise strength is extremely large (e.g., $m$=25 in Table 7), the performance of our HMB algorithm will also be decreased. This is because when the trajectories are extremely noisy and deviated, they will become far different from the standard ones and appear like a different activity. This will obviously affect the recognition performance. However, from Table 7, we can also see that, even in large noise situations, our HMB algorithm can still achieve better performance than the WF-SVM method.

(3) More importantly, note that our HMB algorithm is not limited to trajectories. Instead, various low-level motion features such as the optical flow [28] can also be included into our algorithm to create heat maps for recognition.

Therefore, in cases when reliable trajectories cannot be achieved (such as the $m=25$ case in Table 7), our algorithm can also be extended by skipping the tracking step and directly utilizing other low-level motion features for performing group activity recognition. This point will be further discussed in Section IV-C later.

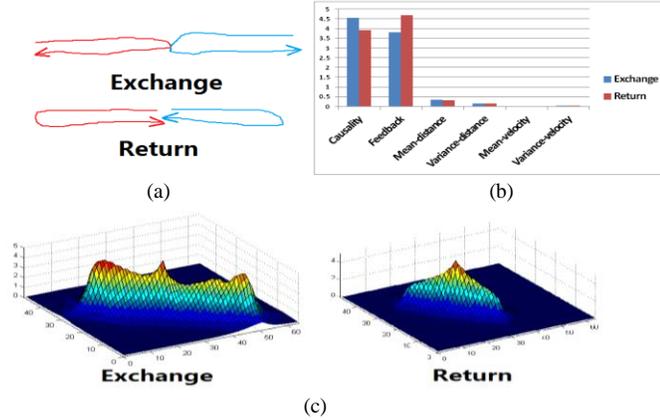

(a)

(b)

(c)

Figure 11. (a) The trajectories for the two complex activities; (b) The major feature values for the WF-SVM algorithm [2]; (c) The HMs for the two complex activities.

Table 8 Miss and TER rates for the complex activities

|  |  | **HMB ($w=1$)** | MF-SVM |
|---|---|---|---|
| Exchange | Miss | **5.9%** | 50.0% |
| Return | Miss | **11.6%** | 43.8% |
| TER |  | **8.8%** | 46.9% |

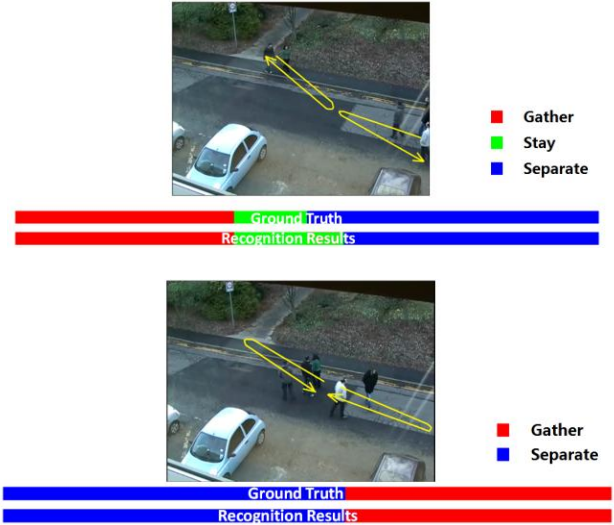

Figure 12. The example frames of the "Exchange" and "Return" sequences and the qualitative results of on-line sub-activity recognition by using a 30-frame-long sliding window. The bars represent labels of each frame, red represents Approach, green represents Stay, and blue represents Separate.

Fourthly, in order to further demonstrate our HM features, we perform another experiment for recognizing two complex activities: "Exchange" (i.e. two people first approach each other, stay together for a while and then separate) and "Return" (i.e., two people first separate and then approach to each other later). In this experiment, we extract 32 pair-trajectories from the BEHAVE dataset for the two complex activities and perform 75% training-25% testing. Some example frames are shown in Figures 11-12. In Figure 11, (a) shows the trajectories of the two complex activities, (b) shows the values of the major features in the WF-SVM algorithm [2], and (c) shows the HM surfaces. From Figure 11 (b), we can see that the features in the WF-SVM algorithm cannot show much difference between the two complex activities. Compared to (b), our HMs in (c) are obviously more distinguishable. The recognition results for the WF-SVM algorithm and our HMB algorithm are shown in Table 8. The results in Table 8 further demonstrate the effectiveness of our HM features in representing complex group activities.

Finally, we evaluate our algorithm in recognizing the sub-activities. Note that our algorithm can be easily extended to recognize the sub-activities by using shorter sliding windows to achieve the short-term trajectories instead of the entire trajectories. By this way, we can also achieve on-the-fly activity recognition at each time instant [6, 29]. In order to demonstrate this point, Figure 12 shows the results by applying a 30-frame-long sliding window to automatically recognize the sub-activities inside the complex "Exchange" and "Return" video sequences. From Figure 12, we can see that our HMB algorithm can also achieve satisfying recognition results for the sub-activities inside the long-term sequences. Besides, our algorithm is also able to recognize both the long-term activities and the short-term activities by simultaneously introducing multiple sliding windows with different lengths. By this way, both the sub-activities of the current clip and the complex activities of the long-term clip can be automatically recognized.

## B. Experimental results for the traffic dataset

In this sub-section, we perform two experiments on the traffic datasets.

Firstly, we perform an experiment on a traffic dataset for recognizing group activities among vehicles in the crossroad. The dataset is constructed from 20 long surveillance videos taken by different cameras. Seven vehicle group activities are defined as in Table 9 and some example activities are shown in Figure 13. We select 245 video clips from the dataset where each activity includes 35 video clips and each clip includes two trajectories. In this dataset, the trajectories are achieved by first using our proposed object detection method [26] to detect the vehicles and then using the particle-filtering-based tracking method [9, 31] to track the detected vehicles. The Miss, FA, and TER of different algorithms are shown in Table 10.

Table 9 Definitions of the vehicle group activities

| Turn | A car goes straight and a car in another lane turns right. |
|---|---|
| Follow | A car is followed by a car in the same lane. |
| Side | Two cars go side-by-side in two lanes. |
| Pass | A car passes the crossroad and a car in the other direction waits for green light. |
| Overtake | A car is overtaken by a car in a different lane. |
| Confront | Two cars in opposite directions go by each other. |
| Bothturn | Two cars in opposite directions turn right at the same time. |

From Table 10, we can see that the LC-SVM algorithm produces less satisfactory results. This is because the group activities in this dataset contain more complicated activities that are not easily distinguishable by the causality and feedback features [3]. Also, the performance of the WF-SVM and the GRAD algorithms are still unsatisfactory in several activities such as "Follow", "Overtake", and "Pass". Compared to these, the performances of our HM algorithms (HMB ($w$=1) and HM-PCASVM) are obviously improved. Besides, the performance of the HMB ($w$=1) is also improved from the HM-PCASVM algorithm. These further demonstrate the effectiveness of our proposed HM feature as well as our SF recognition method.

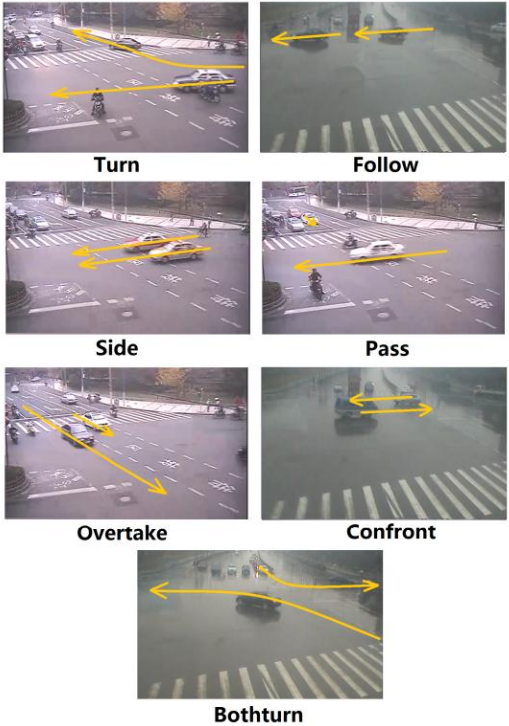

Figure 13. Examples of the defined vehicle group activities.

Table 10 Miss, FA, and TER for different algorithms on the vehicle group activity dataset

|  |  | HMB (w=1) | HM-PCASVM | WF-SVM [2] | LC-SVM [3] | GRAD [6] |
|---|---|---|---|---|---|---|
| Turn | Miss | 2.9% | 4.7% | 2.0% | 16.9% | 10.7% |
|  | FA | 0.5% | 1.4% | 0.5% | 5.4% | 4.1% |
| Follow | Miss | 11.4% | 9.6% | 22.9% | 38.1% | 15.4% |
|  | FA | 0.5% | 1.0% | 4.4% | 15.1% | 5.9% |
| Side | Miss | 1.9% | 0.1% | 0.2% | 16.5% | 7.1% |
|  | FA | 1.0% | 1.9% | 0.3% | 1.0% | 0.3% |
| Pass | Miss | 0.0% | 5.7% | 11.7% | 17.6% | 15.5% |
|  | FA | 0.1% | 0.0% | 2.4% | 1.5% | 3.1% |
| Overtake | Miss | 5.7% | 11.4% | 47.1% | 61.7% | 36.6% |
|  | FA | 0.5% | 0.2% | 4.9% | 11.7% | 3.8% |
| Confront | Miss | 5.6% | 9.5% | 3.9% | 19.6% | 12.4% |
|  | FA | 1.9% | 2.9% | 1.5% | 10.7% | 8.3% |
| Bothturn | Miss | 2.9% | 3.0% | 1.2% | 2.9% | 4.2% |
|  | FA | 1.0% | 1.4% | 0.5% | 1.0% | 2.9% |
| TER |  | 4.5% | 7.8% | 12.1% | 21.5% | 14.6% |

Furthermore, it should be noted that there are two important challenging characteristics for the traffic dataset: (1) The videos in the dataset are taken from different cameras (as in Figure 13). This makes the trajectories vary a lot for the same activity. (2) Within each video, there are also large scale variations (i.e., the object size is much larger at the front region than that in the far region, as shown in Figure 13). Because of this, same activities from different regions may also have large variations and are difficult to be differentiated. These challenging characteristics partially lead to the low performance in the compared algorithms (WF-SVM, LC-SVM, and GRAD). However, comparatively, these variations in scale and camera view are much less obvious in our HM algorithms (HMB ($w$=1) and HM-PCASVM) by utilizing the proposed KPB alignment method for eliminating the scale differences and utilizing the proposed HM for effectively catching the common characteristics of activities.

Secondly, we also perform another experiment with different camera settings. In this experiment, we use the traffic dataset in Figure 13 to train the HMs and then directly use these HMs to recognize the activities from a new dataset as in Figure 14. The new dataset in Figure 14 includes 65 video clips taken from a camera whose height, angle, and zoom are largely different from the ones in Figure 13. The results are shown in Table 11.

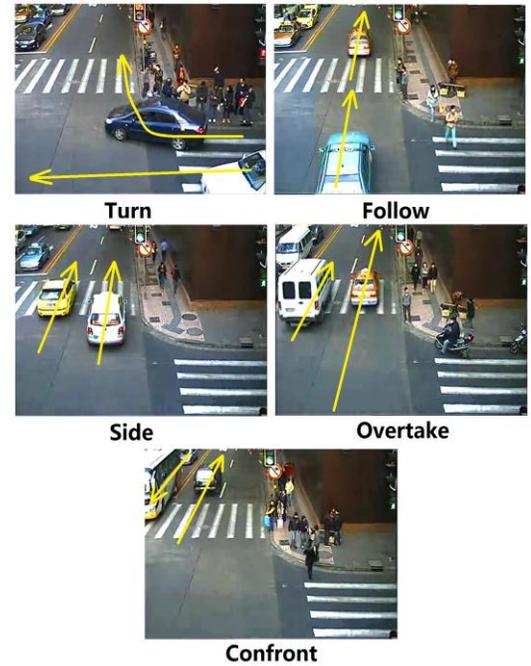

Figure 14. Examples of vehicle group activities in the new dataset.

Table 11 Miss, FA, and TER for different algorithms by using the HMs trained from the traffic dataset in Figure 13 to recognize the new dataset in Figure 14

|  |  | HMB (w=1) | HM-PCASVM | WF-SVM[2] | LC-SVM[5] | GRAD [6] |
|---|---|---|---|---|---|---|
| Turn | Miss | 0.0% | 0.0% | 0.0% | 38.5% | 15.2% |
|  | FA | 2.1% | 3.6% | 0.0% | 13.5% | 4.8% |
| Follow | Miss | 8.3% | 7.7% | 32.1% | 46.2% | 20.1% |
|  | FA | 1.9% | 0.0% | 17.3% | 13.5% | 10.6% |
| Side | Miss | 12.8% | 18.2% | 28.6% | 30.8% | 35.5% |
|  | FA | 4.1% | 1.7% | 0.0% | 0.0% | 0.0% |
| Pass | Miss | 0.0% | 0.0% | 0.0% | 0.0% | 0.0% |
|  | FA | 0.0% | 0.0% | 9.2% | 1.3% | 3.1% |
| Overtake | Miss | 9.7% | 15.1% | 53.8% | 61.5% | 46.9% |
|  | FA | 3.8% | 7.2% | 1.9% | 7.7% | 2.7% |
| Confront | Miss | 6.2% | 8.7% | 23.1% | 46.2% | 16.2% |
|  | FA | 1.0% | 0.0% | 0.0% | 3.8% | 1.3% |
| Bothturn | Miss | 0.0% | 0.0% | 0.0% | 0.0% | 0.0% |
|  | FA | 0.0% | 1.5% | 3.1% | 9.2% | 3.3% |
| TER |  | 6.9% | 10.1% | 27.7% | 44.6% | 32.9% |

From Table 11, we can see that when using the trained models to recognize the activities in a dataset with different camera settings, the performances of the compared algorithms (WF-SVM, LC-SVM, and GRAD) are obviously decreased. Comparatively, our HM algorithms (HMB ($w$=1) and HM-PCASVM) can still produce reliable results. This demonstrates that: (1) Our proposed key-point-based (KPB) heat-map alignment method can effectively handle the heat map differences due to different camera settings. (2) Our HMB algorithm has the flexibility of directly applying the HMs trained from one camera setting to the other camera settings.

## C. Experimental results for the UMN dataset

Finally, in order to demonstrate that our algorithm can also be extended to other low-level motion features [28], we perform another experiment by using the optical flows for recognition.

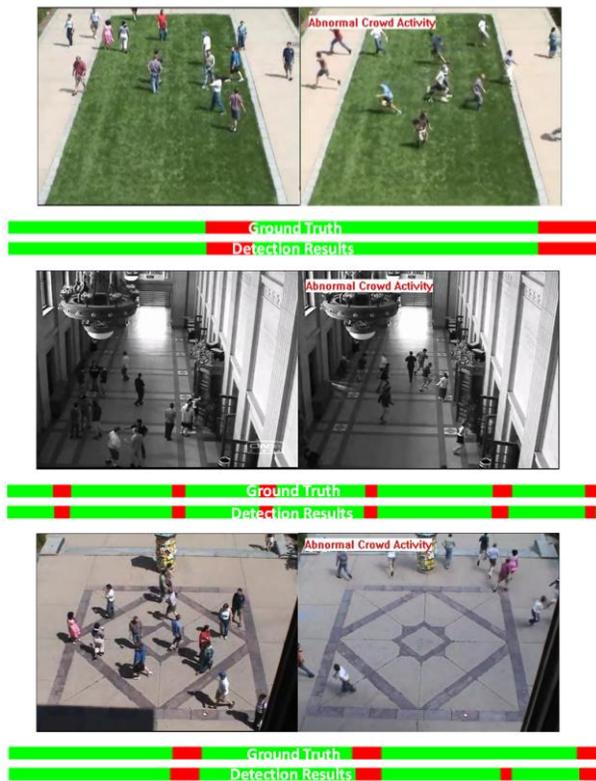

Figure 15. The qualitative results of using our HMB algorithm for abnormal detection in the UMN dataset. The bars represent the labels of each frame, green represents normal and red represents abnormal.

The experiment is performed on the UMN dataset [22] which contains videos of 11 different scenarios of an abnormal escape event in 3 different scenes including both indoor and outdoor. Each video starts with normal behaviors and ends with the abnormal behavior (i.e., escape). In this experiment, we first compute the optical flow between the current and the previous frames. Then patches with high optical-flow magnitudes will be viewed as the heat sources for creating the heat maps (HMs) and these heat maps will be utilized for activity recognition in our HMB algorithm. A sliding window of 30 frames is used as the basic video clip and one HM is generated from each clip. By this way, we can achieve 257 video clips. We randomly select 5 normal behavior HMs and 5 abnormal behavior HMs as the training set to classify the rest 247 video clips. Furthermore, we set $w$ in Eq. (5) as 1.

Figure 15 shows some example frames of the UMN dataset and compares the normal/abnormal classification results of our algorithm with the ground truth. Furthermore, Figure 16 compares the ROC curves between our algorithm (HMB+ Optical Flow) and three other algorithms: the optical flow only method (Optical Flow) [20, 28], the Social Force Model (SFM) [20], and the Velocity-Field Based method (VFB) [21].

From Figure 15 and 16, we can have the following observations:

(1) From Figure 15, we can see that the UMN dataset includes high density of people where reliable tracking is difficult. However, our HMB algorithms can still achieve satisfying normal/abnormal classification results by using the optical flow features. This demonstrates the point that when reliable trajectories cannot be achieved, our algorithm can be extended by skipping the tracking step and directly utilizing the low-level motion features to perform group activity recognition.

(2) From Figure 15, we can see that our algorithm can perform online normal/abnormal activity recognition for each time instant by using a 30-frame-long sliding window. This further demonstrates that our algorithm is extendable to on-the-fly and sub-activity recognitions.

(3) Our HMB algorithm can achieve similar or better results than the existing social-force-based methods [20-21] when detecting the UMN dataset. This demonstrates the effectiveness of our HMB algorithm. Although other social-force-based methods [23] may have further improved results on the UMN dataset, the performance of our HMB algorithm can also be further improved by: (a) using more reliable motion features (such as the trajectories of the local spatio-temporal interest points [23]) to take the place of the optical flow, (b) including more training samples (note that in this experiment, only five normal clips and five abnormal clips are used for training in our algorithm).

(4) More importantly, compared with the social-force-based methods [20, 21, 23], our HMB algorithm also has the following advantages:

(a) Most social-force-based methods [20, 21, 23] are more focused on the relative movements among the objects (e.g., whether two objects are approaching or splitting) while the objects' absolute movements in the scene are neglected (e.g., whether an object is stand still or moving in the scene). Thus, these methods will have limitations in differentiating activities with similar relative movements but different absolute movements (such as "Wait" and "Gather" in Figure 10 or "Confront" and "Bothturn" in Figure 13). Comparatively, the heat map features in our HMB algorithm can effectively embed both the relative and absolute movements of the objects.

(b) Since the interaction forces used in the social-force-based methods [20, 21, 23] cannot effectively reflect the correlation changes over time (e.g., two objects first approach and then split), they also have limitations in differentiating activities with